%% file: main.tex
\documentclass[11pt]{article}

\usepackage{blindtext}

\usepackage[preprint]{jmlr2e}
\usepackage{statistics}
\usepackage{multirow}

\usepackage{algorithm2e}
\usepackage[toc,page]{appendix}
\usepackage{listings}
\usepackage{makecell,multirow,enumitem,subfig,bm,booktabs}
\usepackage{wrapfig}
\usepackage{etoolbox}
\usepackage{url}            
\usepackage{booktabs}       
\usepackage{amsfonts}       
\usepackage{nicefrac}       
\usepackage{microtype}      
\usepackage{xcolor}         
\usepackage{amsmath,amssymb,mathtools}
\usepackage{cleveref}
\usepackage{pifont}
\usepackage{listings}
\usepackage{xcolor}
\usepackage{minitoc}
\usepackage{caption}
\usepackage{xcolor}
\usepackage{inconsolata}
\usepackage[textsize=tiny]{todonotes}

\lstdefinestyle{mypython}{
    language=Python,
    backgroundcolor=\color{gray!10},
    basicstyle=\ttfamily\small,
    keywordstyle=\color{blue},
    stringstyle=\color{red},
    commentstyle=\color{green!50!black},
    showstringspaces=false,
    breaklines=true,
    frame=single
}

\def\wty#1{\textcolor{black}{#1}}

\usepackage{xcolor}

\usepackage{lastpage}
\jmlrheading{23}{2026}{1-\pageref{LastPage}}{1/21; Revised 5/22}{9/22}{21-0000}{TW, JL, HL, HN, JB}

\ShortHeadings{DRO Package}{}
\firstpageno{1}

\usepackage{etoolbox} 
\pretocmd{\section}{\csname phantomsection\endcsname\addtocontents{dummy}{}}{}{}

\begin{document}

\title{
DRO: A Python Library for Distributionally Robust Optimization in Machine Learning}

\author{\name Tianyu Wang$^*$\email tw2837@columbia.edu\\    
\name Jiashuo Liu\thanks{T. Wang and J. Liu contributed equally and are listed in reverse alphabetical order.} \email liujiashuo77@gmail.com\\
        \name Henry Lam  \email khl2114@columbia.edu\\
        \name Hongseok Namkoong \email namkoong@gsb.columbia.edu\\
        \name Jose Blanchet \email jose.blanchet@stanford.edu\\
       }
\editor{My editor}

\maketitle

\begin{abstract}
We introduce \texttt{dro}, an open-source Python library for distributionally robust optimization (DRO) for regression and classification problems. The library implements 15 DRO formulations and 9 backbone models, enabling 85 distinct DRO methods. Furthermore, \texttt{dro} is compatible with both \texttt{scikit-learn} and \texttt{PyTorch} \wty{through a unified interface, and uses approximation techniques to reduce runtime for some computationally demanding formulations}. 
Comprehensive documentation is available at \url{https://python-dro.org}.
\end{abstract}

\vspace{-0.2in}
\section{Introduction}

Robustness is critical in high-stakes application domains of machine learning and decision making, such as finance~\citep{giudici2023safe,blanchet2022distributionally}, healthcare~\citep{wong2021external} and supply chain~\citep{tang2006robust,albahri2023systematic}. A central problem in these settings is the presence of distribution shifts between the training data and deployment environment~\citep{quinonero2008dataset}. 
To address this, Distributionally Robust Optimization (DRO) has emerged as a key framework for training predictive models and making decisions that perform well under worst-case scenarios over a set of plausible deployment distributions~\citep{kuhn2024distributionally}. With various specifications of distributional uncertainty sets~\citep{blanchet2024distributionally,liu2024rethink}, a growing number of DRO formulations have been developed to tackle various types of distribution shifts. 

However, the practical adoption of DRO remains limited because general-purpose computational tools remain limited. A key challenge lies in the intrinsic difficulty in solving DRO problems, which typically involve min-max optimization. The tractability of these problems crucially depends on the structure of the predictive model and loss function~\citep{kuhn2024distributionally}. Even in simple linear settings, where reformulations may allow the use of standard solvers, popular variants of Wasserstein DRO~\citep{esfahani2018data} \wty{often introduce additional variables and constraints beyond the corresponding nonrobust counterpart. These additional computations make high-accuracy solutions costly for some formulations~\citep{wang2021sinkhorn}.}
For widely used neural networks and tree-based ensembles, \wty{tractable reformulations are often unavailable, so training relies on heuristic or approximation algorithms}. 
Correspondingly, existing software tools often fall short in two key aspects:
\begin{enumerate}[noitemsep,leftmargin=*]
    \item[(i)] Scalability and ML integration: Many software tools rely on symbolic reformulations of distributionally robust objectives or constraints to more standard representations and solve the resulting problems using general-purpose solvers~\citep{goh2011robust,chen2020robust,chen2023rsome,vayanos2022roc++}. While helpful, these tools lack encapsulation of the DRO workflow within a unified interface — users must still \wty{formulate the ambiguity set and express the application-specific loss as a worst-case expectation in the tool's modeling language.} Therefore, these tools are difficult to scale or integrate into modern ML workflows. 
    \item[(ii)] Flexibility in modeling: some tools focus on \wty{specific types of} DRO formulations (e.g.,~\cite{vincent2024texttt}), limiting their applicability in handling general distribution shifts. As a result, DRO’s strengths are not yet fully realized in real-world applications, particularly in settings that demand both statistical robustness and computational efficiency.
\end{enumerate}
This paper introduces \texttt{dro}: a library which provides the first comprehensive set of tools by offering modular implementations of different DRO formulations tailored for ML problems, compatible with both \texttt{scikit-learn} and \texttt{PyTorch}. \wty{This package supports 85 methodological combinations across 15 formulations and 9 algorithmic backbones}, encompassing a wide range of formulations including the Wasserstein distance, various $f$-divergences, and group-based subpopulation shifts, and algorithmic backbones including linear, kernel, tree-based, and neural networks with flexible user customization. \wty{Full details are in \Cref{tab:loss_choice}. Note that some formulation-backbone combinations are not implemented yet in the package due to their modeling difficulties, e.g., sample perturbation via Wasserstein DRO under tree-based ensembles requires a different optimization approach}. To reduce the computational cost of MMD-DRO and Marginal-DRO, \wty{we provide approximate formulations that achieve 10--150$\times$ speedups over the unaccelerated implementations without much accuracy degradation in the evaluated settings with details in Appendix~\ref{app:accelerate}}. 
The library provides comprehensive documentation, a modular architecture, and integration with modern ML workflows for researchers and practitioners.

\begin{figure}[t]
    \centering
    \vspace{-0.2in}
    \includegraphics[width=\linewidth]{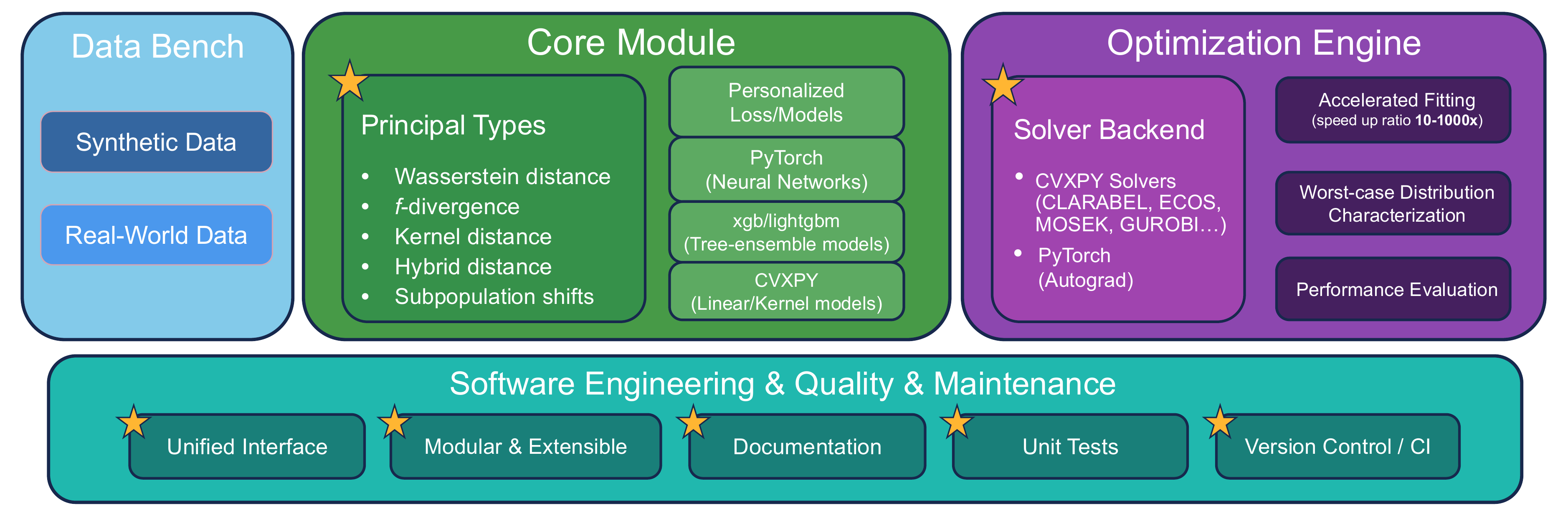}
    \vspace{-0.25in}
    \caption{Overview of the \texttt{dro} library.}
    \label{fig:overview}
    \vspace{-0.2in}
\end{figure}

\vspace{-0.2in}
\section{Package Overview}\label{sec:package}
The \texttt{dro} library is designed to provide an easy-to-use framework for DRO, including data generation, model fitting, and model-based diagnosis, and is built on top of \texttt{CVXPY}~\citep{diamond2016cvxpy} and \texttt{PyTorch}~\citep{paszkepy2019torch}.
The source code of the package is hosted at GitHub (\url{https://github.com/namkoong-lab/dro}) and releases of the \texttt{dro} package are available via PyPI at \url{https://pypi.org/project/dro/}. 
Our documentation is at \url{https://python-dro.org/}, which contains detailed installation instructions, tutorials, examples, and API references. 
An overview of the \texttt{dro} library is shown in~\Cref{fig:overview}, and the following summarizes our supported methods. We defer additional components (data generation and model-based diagnostics, including generating the worst-case distribution and performance evaluation of some DRO methods) to Appendix~\ref{app:dgp} and~\ref{app:diagnostics}, and full modeling details of the 85 methodological combinations to Appendix~\ref{app:dro-model}.
\vspace{-0.1in}
\paragraph{Problem Formulation.}
The \texttt{dro} library addresses supervised learning problems with training samples $\{(x_i,y_i)\}_{i=1}^n$, where $x_i \in \mathcal{X}$ denotes a feature vector and $y_i \in \mathcal{Y}$ denotes the corresponding label. Given a model class $\mathcal{F}$ of predictors $f \colon \mathcal{X} \to \mathcal{Y}$ and a loss function $\ell \colon \mathcal{Y} \times \mathcal{Y} \to \mathbb{R}_+$, the library solves DRO problems as:
\vspace{-0.1in}
\begin{equation}\label{eq:general-dro}
    \min_{f \in \mathcal{F}} \sup_{Q \in \mathcal{P}} \mathbb{E}_Q\big[\ell(f(X), Y)\big],
\vspace{-0.1in}
\end{equation}
where $\mathcal{P}$ denotes the \emph{ambiguity set} containing the empirical distribution and its neighborhood. The objective controls worst-case loss over distributions in the chosen ambiguity set.
A common distance-based ambiguity set $\mathcal{P}(d, \epsilon) = \{ Q : d(Q, \widehat{P}_n) \leq \epsilon \}$ characterizes distributions within $\epsilon$-radius of the empirical distribution $\widehat{P}_n := \frac{1}{n}\sum_{i=1}^n \delta_{(x_i,y_i)}$, where $d(\cdot,\cdot)$ is a probability distance metric (e.g., Wasserstein distance) and $\epsilon > 0$ controls the robustness level. Coverage guarantees require suitable calibration of the ambiguity set, and computational tractability depends on the model and loss structure. Group-based ambiguity sets instead vary the mixture proportions of observed subpopulations, as detailed in Appendix~\ref{app:dro-model}.
The library's DRO formulations adapt to three core components: the \emph{ambiguity-set specification}, \emph{model class} $f(\cdot)$, and \emph{loss function} $\ell$. 

\paragraph{Supported Ambiguity Sets.} 
We support 5 principal types of ambiguity sets and 15 DRO formulations in total: 
\begin{enumerate}[noitemsep,leftmargin=*]
    \item[(1)] \emph{Wasserstein distance}~\citep{blanchet2019data,gao2016distributionally,shafieezadeh2019regularization} via allowing changes in $X$ and (or) $Y$,~\wty{including the robust satisficing variant,  RS-WDRO~\citep{long2023robust}};
    \item[(2)] \emph{$f$-divergence} including KL-divergence (KL)~\citep{hu2013kullback}, $\chi^2$-divergence~\citep{lam2019recovering,duchi2019variance}, Total Variation~\citep{jiang2018risk}, Bayesian variants~\citep{shapiro2023bayesian}, and conditional value-at-risk (CVaR) with variants for marginal $X$ shifts, i.e., Marginal-DRO~\citep{duchi2021learning}, and conditional $Y|X$ shifts~\citep{sahoo2022learning};
    \item[(3)] \emph{Kernel distance} methods~\citep{zhu2021kernel}, i.e., Maximum Mean Discrepancy (MMD)-DRO;
    \item[(4)] \emph{Hybrid distance} measures combining multiple metrics, such as Holistic Robust (HR)-DRO~\citep{bennouna2022holistic}, Sinkhorn distance~\citep{wang2021sinkhorn}, Outlier-robust Wasserstein Distance~\citep{nietert2023outlier}, and Moment Optimal Transport Distance (MOT)~\citep{blanchet2023unifying};
    \item[(5)] \wty{\emph{Subpopulation shift} modeled by Group-DRO~\citep{sagawa2020distributionally}, which optimizes against the group with the worst empirical performance.}
\end{enumerate}
\vspace{-0.15in}
\paragraph{Supported Model Classes \& Loss Functions.} 
The \texttt{dro} library supports linear models, kernel methods, tree methods, and neural networks. \wty{According to their optimization procedures, the supported methods fall into two categories:} 
\begin{enumerate}[noitemsep,leftmargin=*]
    \item[(i)] \textit{Exact optimization} (convex optimization for linear cases: classification via hinge/logistic loss, regression via \wty{LAD or OLS loss} and extensions to kernel methods) for distance-based and group-based methods. To ensure optimization fidelity, each model is fitted through disciplined convex programming using \texttt{CVXPY} backends and vectorization techniques. For two computationally challenging DRO models (MMD-DRO and Marginal-DRO) under large-scale problems, we further conduct 
Nyström approximation and constraint subsampling for MMD-DRO and sparse matrix reformulation for Marginal-DRO \wty{to reduce the size of their respective optimization problems}. These optimization approximations enable efficient \wty{fitting and comparison of supported DRO models on datasets of practical size.} 
Computational details are in Appendix~\ref{app:accelerate}.
    \item[(ii)] \textit{Approximate optimization} methods (tree-based ensembles and neural networks) for classical Wasserstein, $f$-divergence-based, and Group-DRO methods. To ensure efficient deployment, the tree-based ensemble model is fitted on top of \texttt{lightgbm} or \texttt{xgboost} and the neural network model is fitted on top of \texttt{PyTorch}. \wty{The neural Wasserstein implementation supports norm-ball adversarial training using \texttt{Torchattacks}~\citep{kim2020torchattacks} and a transport-penalized inner search following \citet{Sinha2017certifying}. The neural $f$-divergence methods build on the framework of \citet{levy2020large}.}
\end{enumerate}
\textbf{Personalization.} For each of the optimization procedures above, our framework further accommodates user-defined loss functions and model architectures, 
particularly for Wasserstein distance and standard $f$-divergence ambiguity sets through a compatible abstraction via \wty{\texttt{CVXPY} and \texttt{PyTorch} modules}. The available personalization hooks depend on the model class and loss function (See Appendix~\ref{app:personalization} for more details). 

\vspace{-0.15in}
\section{Modular Design Merits}
In this section, we further highlight two modular design merits.
\vspace{-0.1in}
\paragraph{Standard Interface.}
All methods provide a consistent API following \texttt{scikit-learn}-like estimator interface, including standard functions such as \texttt{update()}, \texttt{fit()}, \texttt{predict()}, and \texttt{score()}. 
This design ensures full compatibility with \texttt{scikit-learn}'s ecosystem while maintaining each method's specific parameters through inheritance. 
The interface strictly follows \texttt{scikit-learn}'s API guidelines to guarantee seamless integration with existing machine learning workflows. We provide a demo example in Appendix~\ref{app:demo}.
\vspace{-0.1in}
\paragraph{Software Engineering Compliance.}
The package adheres to rigorous software engineering standards, featuring:
(1) comprehensive unit testing with 91\% line coverage across core algorithms;
(2) static type checking via Python's MyPy at compile-time; (3) continuous integration testing across Python 3.8+ environments (Linux/Windows/macOS).
These practices ensure the implementation maintains both precision and production reliability.

\vspace{-0.15in}
\section{Conclusion}
Currently, the \texttt{dro v.0.4.3} library provides the most comprehensive implementation of data-driven DRO methods for supervised learning, supporting both diverse distance metrics and model architectures, \wty{as shown by the comparison with existing software in Appendix~\ref{app:compare}}. In the future, we plan to include  efficient support for \wty{sequential decision-making or multi-stage DRO~\citep{bertsimas2019adaptive} and DRO under multiple ambiguity sets~\citep{selvi24,wang2024contextual,rychener2024wasserstein}}.

\newpage
\vskip 0.2in
\bibliography{cite}

\newpage
\appendix
\clearpage



\section{Full Details of DRO Models}\label{app:dro-model}
Here, we specify details of each ambiguity set of the general DRO problem in~\eqref{eq:general-dro} with the ambiguity set $\Pscr$ (or $\Pscr(d,\epsilon)$ more specifically). We refer readers to the complete modeling details in \url{https://python-dro.org/api/tutorials.html}. Complete method specifications are detailed in \Cref{tab:loss_choice}.

\begin{table}[htb]
    \caption{Different DRO methods supported in \texttt{dro} package.}
    \label{tab:loss_choice}
    \small
    \resizebox{\textwidth}{!}{
    \begin{tabular}{l|cccccc|ccc}
        \toprule
         & \multicolumn{6}{c|}{Exact Optimization} & \multicolumn{3}{c}{Approximate Optimization} \\
         & LAD & OLS & SVM & Logistic & Kernel & Personal & Tree-based & NN & Personal\\
         \midrule
        WDRO & \cmark & \cmark & \cmark & \cmark & \cmark & \cmark & & \cmark & \cmark \\
        RS-WDRO & \cmark & \cmark & \cmark & \cmark & \cmark & \cmark &\\
        \midrule
        $\chi^2$-DRO & \cmark & \cmark & \cmark & \cmark & \cmark & \cmark & \cmark &\cmark&\cmark\\
        KL-DRO & \cmark & \cmark & \cmark & \cmark & \cmark & \cmark & \cmark &\\
        Bayesian-DRO & \cmark & \cmark & \cmark & \cmark & & & & &\\
        CVaR-DRO & \cmark & \cmark & \cmark & \cmark & \cmark & \cmark & \cmark & \cmark & \cmark\\
        TV-DRO & \cmark & \cmark & \cmark & \cmark & \cmark & \cmark &\\
        Marginal(-CVaR)-DRO & \cmark & \cmark & \cmark & \cmark & \cmark & \cmark &\\
        Conditional(-CVaR)-DRO & \cmark & \cmark & \cmark & \cmark & \cmark & \cmark &\\
        \midrule
        MMD-DRO& \cmark & \cmark & \cmark & \cmark & & \cmark \\
        \midrule
        HR-DRO &\cmark & & \cmark & & & & & \cmark& \cmark \\
        Sinkhorn-DRO & \cmark & \cmark & \cmark & \cmark & & & &\cmark &\\
        OutlierRobust(OR)-WDRO & \cmark & & \cmark & & & & & \\
        MOT-DRO &\cmark & & \cmark & & & & & \\
        \midrule
        Group-DRO & \cmark & \cmark & \cmark & \cmark & \cmark & & & \cmark & \\
        \bottomrule
    \end{tabular}
    }
    \vspace{0.15em}
    {\wty{\emph{Notes.} A checkmark indicates that the package implements the indicated DRO formulation (in the row) for the model or loss family (in the column) under the supported settings documented. It does not imply support for every possible instance in that family.}}
\end{table}

\paragraph{\bf (1) Wasserstein Distance.}~In the standard Wasserstein-DRO (\textbf{WDRO}), we apply $d(P, Q) = W(P, Q)$, where $W(\cdot,\cdot)$ is a Wasserstein optimal-transport discrepancy:
\begin{equation*}
W(P_1, P_2) = \inf_{\pi \sim (P_1, P_2)}\mathbb{E}_{\pi}[c(Z_1, Z_2)],
\end{equation*}
where $Z_1 = (X_1, Y_1), Z_2 = (X_2, Y_2)$, and $c:\mathcal Z\times\mathcal Z\to[0,\infty]$ is the ground transportation cost on $\mathcal Z=\mathcal X\times\mathcal Y$. The notation $\pi\sim(P_1,P_2)$ means that $\pi$ is a coupling with marginals $P_1$ and $P_2$; we denote the set of such couplings by $\Pi(P_1,P_2)$. For a ground metric $\rho$ on $\mathcal Z$ and $1\leq p<\infty$, the $p$-Wasserstein distance is
\[
W_p(P_1,P_2)=\left(\inf_{\pi\in\Pi(P_1,P_2)}\mathbb E_{\pi}\bigl[\rho(Z_1,Z_2)^p\bigr]\right)^{1/p}.
\]
Thus, choosing $c(z_1,z_2)=\rho(z_1,z_2)^p$ in the preceding display gives $W(P_1,P_2)=W_p(P_1,P_2)^p$. The Wasserstein order and the feature-norm order specified below are independent choices.

In the linear and kernel models, we specify the cost metric for each loss type: (a) in ``lad'', \wty{we use the 1-Wasserstein distance, and the ground cost is} $c((X_1, Y_1), (X_2, Y_2)) = \|(X_1 - X_2, Y_1 - Y_2)\|$; (b) in ``ols'', \wty{we use squared 2-Wasserstein feature transport with fixed targets: $c((X_1,Y_1),(X_2,Y_2))=\|X_1-X_2\|_{\Sigma,p}^2$ when $Y_1=Y_2$, and $+\infty$ otherwise. For LAD, the weighted product norm determined by $(\Sigma, p, \kappa)$} on the product space $\mathcal{X} \times \mathbb{R}$ is:
\[\|(x, y)\| = \|x\|_{\Sigma, p} + \kappa |y|.\]
\wty{(c) in binary classification (``svm'' and ``logistic''), we instead use the discrete-label ground cost $c((X_1,Y_1),(X_2,Y_2))=\|X_1-X_2\|_{\Sigma,p}+\kappa\mathbf{1}\{Y_1\neq Y_2\}$, so that every label mismatch costs exactly $\kappa$.} Here $\|x\|_{\Sigma, p} = \|\Sigma^{1/2}x\|_p$. $\Sigma$ is the identity matrix in the default setup and $\kappa$ denotes the robustness parameter for the perturbation $Y$; \wty{setting $\kappa = \infty$ forbids perturbations of $Y$. The implementation requires this setting for OLS, requires $\kappa>0$ for LAD, and permits $\kappa=0$ for free binary-label flips.} The parameter $p$ controls the perturbation norm of $X$.

\wty{For neural Wasserstein DRO, the \texttt{norm\_ball} mode uses fixed-radius, label-preserving attacks from \texttt{Torchattacks}~\citep{kim2020torchattacks}, with a $\infty$-Wasserstein interpretation under the corresponding norm-ball perturbation assumptions. The \texttt{transport\_penalty} mode follows \citet{Sinha2017certifying} and approximately optimizes a fixed-penalty squared-transport objective associated with 2-Wasserstein DRO; it does not exactly solve a DRO problem with a prescribed radius. These implementations build on the adversarial learning literature~\citep{yuan2019adversarial}. For broader comparisons, we refer readers to \citet{croce2021robustbench,li2024oodrobustbench}.}

For the {Robust Satisficing Wasserstein-DRO} (\textbf{RS-WDRO}) method \citep{long2023robust}, we fix a target loss $\tau$ and minimize the sensitivity coefficient $\lambda$:
\[
\begin{aligned}
\min_{\theta,b,\lambda\geq0}\quad &\lambda\\
\text{s.t.}\quad &\mathbb E_P[\ell_{tr}(\theta,b;(X,Y))]
\leq\tau+\lambda W_c(P,\widehat P),\qquad \forall P.
\end{aligned}
\]
Smaller $\lambda$ gives a tighter bound on loss growth under distribution shift. For linear predictors with hinge, logistic, or absolute losses and label-preserving norm transport, this reduces to
\[
\min_{\theta,b}\;\|\Sigma^{-1/2}\theta\|_q
\quad\text{s.t.}\quad
\mathbb E_{\widehat P}[\ell_{tr}(\theta,b;(X,Y))]\leq\tau,
\qquad \frac1p+\frac1q=1,
\]
where $q$ is the dual of the feature perturbation norm order $p$. For LAD with finite label transport cost $\kappa$, the implementation instead minimizes $\max\{\|\Sigma^{-1/2}\theta\|_q,1/\kappa\}$ under the same empirical-loss constraint.
The implementation specifies the loss target through a \emph{target ratio} $\beta\geq1$, with $\tau=\beta\,\mathbb E_{\widehat P}[\ell_{tr}(\widehat\theta,\widehat b;(X,Y))]$, where $(\widehat\theta,\widehat b)$ is an empirical risk minimizer. The constraint $\beta\geq1$ applies to this multiplier, not to the loss target $\tau$.

\paragraph{\bf (2) (Generalized) $f$-divergence.} When $d$ is set as the generalized $f$-divergence (including CVaR), all distances there can be formulated as follows:
\[d(P, Q) = E_{Q}\Paran{f\Para{\frac{dP}{dQ}}}.\] 

For the \textbf{KL-DRO} method \citep{hu2013kullback}, we apply $f(x) = x \log x - (x - 1)$;

For the \textbf{$\chi^2$-DRO} method \citep{duchi2019variance},  we apply $f(x) = (x - 1)^2$;

For the \textbf{TV-DRO} method \citep{jiang2018risk}, we apply $f(x) = \frac12|x - 1|$, giving $d(P,Q)=\frac12\int|dP-dQ|$;

In each of the three methods above, the hyperparameter is $\epsilon$ as the uncertainty set size.
For TV-DRO, the constructor accepts $0\leq\epsilon<1$, while parameter updates require $0<\epsilon<1$; zero radius gives empirical risk minimization.

For the (joint) \textbf{CVaR-DRO} problem \citep{rockafellar2000optimization},
we apply $f(x) = 0$ if $x \in [0, 1/\alpha]$ and $\infty$ otherwise (which is a generalized $f$-divergence). Here the hyperparameter $\alpha$ denotes the worst-case ratio. And we denote such induced $d$ as the so-called \emph{CVaR distance}. 

\textbf{Bayesian DRO}~\citep{shapiro2023bayesian} motivates the following nested objective based on a distribution prior:
\[\min_{\theta \in \Theta}\mathbb{E}_{\zeta \sim \zeta_N}[\sup_{Q \in \mathcal{Q}_{\zeta}(d, \epsilon)} \mathbb{E}_{\xi \sim Q}[\ell(\theta;\xi)]],\]
where $\zeta_N$ denotes the posterior distribution of the parameter $\zeta$ given $\{\xi_i\}_{i \in [N]}$. The ambiguity set is $\mathcal{Q}_{\zeta}(d,\epsilon)=\{Q:d(Q,P_{\zeta})\leq\epsilon\}$, where $P_{\zeta}$ denotes the distribution parametrized by $\zeta$.

In practice, we use finitely many posterior draws and Monte Carlo samples from each corresponding nominal distribution. The $\chi^2$ branch uses separate dual variables for each posterior draw, whereas the KL branch shares one nonnegative dual temperature across draws, giving a restricted-dual upper bound on the sum of separately robustified risks. Both branches minimize the sum over posterior draws.

Besides these standard (generalized) $f$-divergence DRO models, we also include DRO models modeling partial distribution shifts on $(X, Y)$ based on CVaR metrics. Here, we directly use $\mathcal{P}(\alpha)$ as the ambiguity set where $\alpha$ is the CVaR parameter.

If we only consider the shifts in the marginal distribution $X$, we formulate the \textbf{Marginal-CVaR-DRO} model:
\[\mathcal{P}(\alpha) = \{Q_0: P_X = \alpha Q_0 + (1-\alpha) Q_1, \text{for some}~\alpha \geq \alpha_0~\text{and distribution}~Q_1~\text{and}~\mathcal{X}\}.\]
Specifically, we follow the formulation of (27) in \cite{duchi2023distributionally} to fit the model.

If we consider the shift in the conditional distribution $Y|X$,  we formulate the \textbf{Conditional-CVaR-DRO} model:
\[\mathcal{P}(\alpha) = \{Q_0: P_{Y|X} = \alpha Q_0 + (1-\alpha)Q_1, \text{for some}~\alpha \geq \alpha_0~\text{and distribution}~Q_1~\text{and}~\mathcal{Y}\}.\] 
Specifically, we follow the formulation of Theorem 2 in \cite{sahoo2022learning} to fit the model using the approximation $\alpha(x) = \theta^{\top}x$. 

\paragraph{\bf (3) Kernel distance.}  We include Maximum Mean Discrepancy DRO (\textbf{MMD-DRO}) model with the ambiguity set $\Pscr(d,\epsilon)$, where $d(P, Q) = \|\mu_P - \mu_Q\|_{\mathcal H}$ is the kernel distance with the Gaussian kernel and defined as:
\[\|\mu_P - \mu_Q\|_{\mathcal H}^2 = \mathbb E_{x, x' \sim P}[k(x, x')] + \mathbb E_{y, y' \sim Q}[k(y, y')] - 2\mathbb{E}_{x \sim P, y \sim Q}[k(x, y)], \]
with $k(x, y) = \exp(-\|x - y\|_2^2 / (2\sigma^2))$. In the computation, we apply Equation (7) of Section 3.1.1 in \cite{zhu2021kernel}.

\paragraph{\bf (4) Hybrid distance.} We include other ambiguity set design $\Pscr$ induced by some hybrid distances involved in the ambiguity set $\Pscr$:
\begin{itemize}[leftmargin=*]
    \item \textbf{Sinkhorn-DRO}: $\mathcal{P}(W_{\epsilon};{\rho,\epsilon})= \{P: W_{\epsilon}(P,\hat{P})\leq \rho \}$. Here $W_{\epsilon}(\cdot,\cdot)$ denotes the Sinkhorn Distance, defined as:
\[W_{\epsilon}(P,Q) = \inf_{\gamma \in \Pi(P,Q)}\mathbb{E}_{(x,y)\sim \gamma}[c(x,y)]+\epsilon\cdot H(\gamma\vert \mu\otimes\nu),\]
where $c(x, y) = \|x - y\|_2$ and  $\mu,\nu$ are reference measures satisfying $P\ll \mu$ and $Q\ll \nu$.
    \item  \textbf{Holistic-DRO}: $\mathcal{P}(LP_{\mathcal N}, D_{KL}; \alpha, r) = \{P: P,Q\in\mathcal{P}, LP_{\mathcal N}(\hat{P},Q)\leq \alpha, D_{KL}(Q\|P)\leq r \}$, where $LP(\cdot,\cdot)$ is the Levy-Prokhorov metric $LP_{\mathcal N}(P,Q) = \inf_{\gamma\in\Pi(P,Q)} \mathbb{I}(\xi-\xi'\notin \mathcal{N})d\gamma(\xi, \xi')$ with $\mathcal N$ denoting the perturbed (inaccuracy) set of each sample; $D_{KL}(\cdot\|\cdot)$ is the KL-divergence $D_{KL}(Q\|P) = \int_Q \log \frac{dQ}{dP}dQ$. In our setup, we set the perturbed (inaccuracy) set as a ball $\mathcal N = B_2(0, \epsilon) \times \{0\}$ with parameter $\epsilon$.
    \item \textbf{MOT-DRO}: We implement MOT-DRO using the ambiguity set $\mathcal{P}(M_c;\epsilon) = \{(Q, \delta): M_c((Q, \delta), \tilde P) \leq \epsilon\}$,
where $M_c$ is an optimal-transport discrepancy with moment constraints, defined as:
\[M_c(P,Q)= \inf_\pi \mathbb{E}_\pi[c((Z,W),(\hat Z, \hat W))],\]
where $\pi_{(Z,W)}=P, \pi_{(\hat Z, \hat W)}=Q$, and $\mathbb{E}_\pi[W]=1$. Taking the cost function as
\[c((z,w), (\hat z, \hat w))=\theta_1\cdot w \cdot \|\hat x - x\|_p^s +\theta_2\cdot (\phi(w)-\phi(\hat w))_+,\]
where $\tilde{P} =\hat{P} \otimes \delta_1$ and $z=(x,y)$. The implementation keeps targets fixed, with feature-norm order $p\in\{1,2,\infty\}$ configured by \texttt{p}, transport exponent $s\in\{1,2\}$ by \texttt{square}, and radius $\epsilon$ by \texttt{eps}. Both penalty coefficients default to $1$; updating either to a value greater than $1$ sets the other to satisfy $1/\theta_1+1/\theta_2=1$.
    \item \textbf{OutlierRobust-WDRO}: We implement with the ambiguity set $\mathcal{P}(W_p^{\eta};\epsilon) = \{Q: W_p^{\eta}(Q, \hat P)\leq \epsilon\}$, where:
\[W_p^{\eta}(P, Q) = \inf_{Q' \in \mathcal{P}(R^d), \|Q - Q'\|_{TV}\leq \eta} W_p(P, Q'), \]
where $W_p$ denotes the $p$-Wasserstein distance and $\eta \in [0, 0.5)$ denotes the corruption ratio.
\end{itemize}

\wty{\paragraph{\bf (5) Subpopulation Shift.} We include Group Distributionally Robust Optimization (\textbf{Group-DRO})~\citep{sagawa2020distributionally} to model shifts in the proportions of a finite set of observed groups. Let $g_i \in \mathcal{G}$ denote the group label for sample $(x_i,y_i)$, let $I_g = \{i \in [n]: g_i = g\}$ and $n_g = |I_g|$, and define the empirical distribution within group $g$ as $\widehat P_g = \frac{1}{n_g}\sum_{i \in I_g}\delta_{(x_i,y_i)}$. The Group-DRO ambiguity set contains every mixture of the empirical group distributions:
\[
\mathcal{P}_{\mathrm{group}}
= \left\{Q_q = \sum_{g \in \mathcal{G}}q_g\widehat P_g:
q \in \Delta_{|\mathcal{G}|}\right\},
\]
where $\Delta_{|\mathcal{G}|}$ denotes the probability simplex over the observed groups. Therefore, the general DRO problem in~\eqref{eq:general-dro} becomes
\[
\min_{f \in \mathcal{F}}\sup_{Q \in \mathcal{P}_{\mathrm{group}}}
\mathbb{E}_{Q}[\ell(f(X),Y)]
= \min_{f \in \mathcal{F}}\max_{g \in \mathcal{G}}
\frac{1}{n_g}\sum_{i \in I_g}\ell(f(x_i),y_i).
\]
Unlike distance-ball ambiguity sets, Group-DRO does not require a robustness radius: the observed groups define the adversary's feasible distributions. For linear and kernelized predictors, we solve the exact convex epigraph reformulation. For neural network models, we approximately solve the corresponding min--max problem by applying gradient descent to the model parameters and exponentiated-gradient ascent to the adversarial group probabilities.}

\section{Other Module Components}
In this section, we demonstrate in detail the other components of our \texttt{dro} package.
\subsection{Data Generation}\label{app:dgp}
In addition to the core module and optimization engine, our \texttt{dro} library also supports various synthetic data generating mechanisms that are widely used in the DRO literature, which we refer to as the ``Data Bench'' (see~\Cref{fig:overview}).

\paragraph{Synthetic Classification Datasets.}
We implement four synthetic classification datasets to evaluate model robustness under different structural challenges, inspired by recent DRO literature. 
Each dataset construction is encapsulated in a standalone function, and their origins are summarized in~\Cref{tab:dataset_summary}.

\begin{table}[h]
\centering
\caption{Summary of synthetic classification datasets and their sources.}
\label{tab:dataset_summary}
\resizebox{\textwidth}{!}{
\begin{tabular}{lll}
\toprule
\textbf{Function} & \textbf{Source} & \textbf{Description} \\
\midrule
\texttt{classification\_basic} & Custom & Multi-class Gaussian blobs on a sphere; baseline data generator. \\
\texttt{classification\_DN21} & \cite{duchi2021learning} (Sec 3.1.1) & Linear decision boundary with controlled label noise. \\
\texttt{classification\_SNVD20} & \cite{Sinha2017certifying} (Sec 5.1) & Ring-shaped classification with uncertain margin regions. \\
\texttt{classification\_LWLC} & \cite{Liu2022distributionally} (Sec 4.1) & High-dimensional mixed features with geometric scrambling. \\
\bottomrule
\end{tabular}
}
\end{table}

All generators include a visualization option via a shared utility \texttt{draw\_classification}, and support reproducibility through random seed control.

\paragraph{Regression Datasets.}
We provide five regression data utilities covering synthetic settings and a preprocessed real-world dataset, with a range of structural challenges inspired by recent DRO literature.
The data generation functions and their sources are summarized in~\Cref{tab:regression_dataset_summary}.

\begin{table}[h]
\centering
\caption{Summary of regression datasets and their sources.}
\label{tab:regression_dataset_summary}
\resizebox{\textwidth}{!}{\begin{tabular}{lll}
\toprule
\textbf{Function} & \textbf{Source} & \textbf{Description} \\
\midrule
\texttt{regression\_basic} & Custom & Standard linear regression with Gaussian noise; serves as a baseline. \\
\texttt{regression\_DN20\_1} & \cite{duchi2021learning} (Sec 3.1.2) & Linear model with heteroscedastic noise based on covariate threshold. \\
\texttt{regression\_DN20\_2} & \cite{duchi2021learning} (Sec 3.1.3) & Mixture of linear models simulating group shifts. \\
\texttt{regression\_DN20\_3} & \cite{duchi2021learning} (Sec 3.3) & Real-world UCI crime dataset with pre-processing. \\
\texttt{regression\_LWLC} & \cite{Liu2022distributionally} (Sec 4.1) & Complex high-dimensional regression with group imbalance and spurious correlation. \\
\bottomrule
\end{tabular}
}
\end{table}

Each function returns covariates and targets as \texttt{NumPy} arrays. The generated datasets simulate challenging distribution shifts for robust model evaluation.

\paragraph{Real-World Datasets.}
Real-world datasets can be easily imported using existing Python libraries such as \texttt{whyshift}~\citep{liu2023need} and \texttt{tableshift}~\citep{gardner2023tableshift}.

\subsection{Model-based Diagnostics}\label{app:diagnostics}
To understand the robustness of different methods, we provide two functions given samples post training a model $\hat f$ in the linear modules: (i) Generating Worst-case Distribution: Computing the worst-case distribution (i.e., $Q$ that maximizes $\E_{Q}[\ell(\hat f(X); Y)]$) in Wasserstein distance (based on the worst-case distribution computation in \cite{shafieezadeh2019regularization} and \cite{shafiee2026nash}) or $f$-divergence (based on a direct inner maximization); (ii) Evaluating Out-of-sample Model Performance: Estimate the out-of-sample loss $\E_{P}[\ell(\hat f(X); Y)]$ based on training samples $\{(x_i, y_i)\}_{i \in [n]}$ by correcting in-sample bias in standard $f$-divergence-based DRO models with linear-kernel OLS or logistic loss (based on the bias results in \cite{iyengar2023optimizer}). The inherited \texttt{evaluate()} method is not implemented for SVM, LAD, or nonlinear kernels and raises \texttt{NotImplementedError}; \texttt{score()} provides standard predictive metrics. Furthermore, the methods in our package also support the hyperparameter tuning in \texttt{scikit-learn}.

\subsection{Loss Personalization}\label{app:personalization}
\paragraph{Linear.} When fitting the exact linear models,  users only need to rewrite the corresponding \texttt{self.\_loss()} and \texttt{self.\_cvx\_loss()} functions to change the corresponding losses in the training procedure when creating a new problem instance. This personalization is useful for a variety of $f$-divergence-based DRO (including KL-divergence, $\chi^2$-divergence, CVaR, TV-DRO). For Wasserstein DRO, besides modifying these two loss functions, we also need to modify the \texttt{self.\_penalization()} function to adjust the regularization component, where the regularization component denotes the additional part besides the empirical objective in the Wasserstein DRO objective after the problem reformulation. 

\paragraph{Tree-based Ensembles.} For tree-based ensembles, \texttt{loss(preds, labels)} defines the per-example losses used for robust sample weighting and the reported objective. If a personalized loss changes the prediction derivatives, users must also update the gradients and Hessians supplied by the booster objective, such as \texttt{\_kl\_dro\_loss()} or \texttt{\_cvar\_dro\_loss()}. This applies to the supported KL-, CVaR-, and $\chi^2$-DRO methods using XGBoost or LightGBM.

\wty{\paragraph{Neural Networks.} The preset architectures, selected through \texttt{model\_type}, are a linear network, an MLP, AlexNet, and ResNet-18. By default, training uses an internal training--validation split and the Adam optimizer. We support customization through model/loss and data/optimizer interfaces:
\begin{itemize}[leftmargin=*]
    \item \textbf{Model and Loss Interface.} Across Wasserstein-, HR-, $\chi^2$- and CVaR-DRO, users may replace these presets with a compatible PyTorch \texttt{nn.Module} through \texttt{update\_user\_mode(input\_dim, num\_classes, model, task\_type, device)}. The module must accept the chosen input batches and produce the outputs expected by the task loss (class logits for classification). The personalized loss is implemented differently across different DRO methods: 
     \begin{itemize}[leftmargin=*]
        \item $f$-DRO: users customize the per-example loss by overriding \texttt{\_compute\_individual\_loss()} and passing that robust loss through the model's \texttt{\_criterion()} hook.
        \item Wasserstein DRO: users customize the per-example loss by overriding \texttt{\_loss()}. The \texttt{transport\_penalty} mode uses this loss in both the inner perturbation search and the outer model update. In \texttt{norm\_ball} mode, changing \texttt{\_loss()} changes the outer objective, while the built-in attack continues to use classification cross-entropy unless a matching custom attack is supplied.
        \item HR-DRO: users customize the loss by updating the per-example losses supplied to its robust weighting problem and, when appropriate, its attack objective.
     \end{itemize} 
    \item \textbf{Data and Optimizer Interface.} These neural models accept either \texttt{fit(X, y)}, which splits arrays or tensors into training and validation sets and uses Adam by default, or paired user-provided \texttt{train\_loader} and \texttt{val\_loader} arguments. Users may pass a PyTorch \texttt{optimizer} built from the current network and a learning-rate \texttt{scheduler} that can be stepped after each epoch or training batch. A scheduler requires an explicit optimizer, which should be constructed after installing a custom network. The personalization documentation and notebook provide examples of the loss and training hooks.
\end{itemize}}

\section{Acceleration Details}\label{app:accelerate}
This section provides a detailed overview of the acceleration strategies employed in our \texttt{dro} package, including kernel approximation, and formulation-specific constraint reductions designed to improve computational efficiency across different DRO variants that are implemented in CVXPY.

For the computationally demanding MMD-DRO and Marginal-DRO formulations, we use kernel approximation and formulation-specific constraint reductions to obtain \emph{accelerated} versions. \wty{These constructions change the optimization problem rather than providing an equivalent reformulation.} We refer to our CVXPY implementations of the original formulations as the \emph{unaccelerated} versions, as described below.

\subsection{MMD-DRO}
We implement the original MMD-DRO formulation described by \citet{zhu2021kernel}, whose dense kernel representation involves an $n\times n$ kernel matrix and requires $O(n^2)$ storage and $O(n^3)$ time for its matrix decomposition.
\paragraph{Parallel Kernel Approximation.} 
To reduce the kernel mapping complexity, we apply the \texttt{Nystroem} transformer from \texttt{scikit-learn} to approximate the kernel mapping on the sampled support. The \texttt{n\_components} setting controls the feature dimension; when omitted, it defaults to the smaller of 100 and the support size. Setting \texttt{accelerate=False} uses the full kernel matrix instead.
The full-kernel regularizer is $\eta\sqrt{a^\top K a}$, where $\eta=\epsilon$ is the MMD radius configured by \texttt{eta}. For $K\approx\Phi\Phi^\top$, accelerated fitting defaults to \texttt{accelerated\_regularizer="feature\_coefficients"}, using $\eta\|\beta\|_2$ for the feature coefficients $\beta$, the corresponding approximate RKHS norm. The legacy \texttt{support\_values} option instead penalizes $\eta\|\Phi\beta\|_2$ and remains available explicitly.
The kernel computation logic is summarized as follows:

\begin{lstlisting}[language=Python]
# Kernel computation on sampled support zeta
if accelerate:
    n_components = min(self.n_components or 100, len(zeta))
    nystroem = Nystroem(kernel=self.kernel,
                        n_components=n_components,
                        **self._kernel_parameters(zeta))
    nystroem.fit(zeta)
else:
    K = self._kernel_matrix(zeta)
\end{lstlisting}
Thus, Nyström is selected by \texttt{accelerate=True}, while \texttt{n\_components} sets its rank.

To ensure scalability of MMD-DRO across large datasets and high-dimensional settings, we further parallelize the kernel approximation.
Given that kernel transformations remain the bottleneck in high-dimensional DRO models, we batch the data and apply the Nyström approximation in parallel across multiple CPU cores using \texttt{joblib}:
\begin{lstlisting}[language=Python]
batches = [zeta[i:i+5000] for i in range(0, len(zeta), 5000)]
K_approx_list = Parallel(n_jobs=4)(
    delayed(nystroem.transform)(batch) for batch in batches)
\end{lstlisting}
The transformer is fitted once on \texttt{zeta} before these parallel batch transformations.

\paragraph{Constraint Subsampling.} To further reduce the problem size passed to the convex solver, we randomly select at most 5,000 points from the sampled support for the loss constraints. All loss expressions and constraint terms are computed in a vectorized form:
\begin{lstlisting}[language=Python]
losses = compute_loss(X_selected, y_selected, theta, b)
rhs = f0 + K_approx[selected_indices] @ a
constraints = [losses <= rhs]
\end{lstlisting}
This avoids Python loops and allows \texttt{CVXPY} to efficiently compile the computational graph.

\subsection{Marginal-DRO}
The original formulation of Marginal-DRO in \cite{duchi2023distributionally} involves a full $n \times n$ coupling matrix $B \in \mathbb{R}^{n \times n}_{+}$ over all pairs of training samples, requiring $\mathcal{O}(n^2)$ storage for the coupling variable. Specifically, the loss adjustment term takes the form:
\[
s_i \geq \ell_i - \frac{1}{n} \left( \sum_{j=1}^n B_{ij} - \sum_{j=1}^n B_{ji} \right) - \eta \quad \text{for } i = 1, \dots, n.
\]
This dense matrix variable is computationally prohibitive for moderate-scale datasets, with solver time dominated by the evaluation of the \texttt{CVXPY} graph and the optimization of tens of thousands of primal and dual variables.

We replace the full matrix $B$ with two independent nonnegative auxiliary vectors $B^{\mathrm{row}} \in \mathbb{R}^n$ and $B^{\mathrm{col}} \in \mathbb{R}^n$, respectively. The pairwise distance information is encoded sparsely via a $k$-nearest neighbor (k-NN) graph over the control features:
\[
s_i \geq \ell_i - \frac{1}{n} \left( B^{\mathrm{row}}_i - B^{\mathrm{col}}_i \right) - \eta,
\]
\[
\text{Cost} \propto \sum_{(i,j) \in \text{kNN}} d_{ij} \cdot \frac{B^{\mathrm{row}}_i + B^{\mathrm{col}}_j}{2},
\]
where $d_{ij}$ is a sparse upper-triangular matrix of local distances. For fixed $k$, this surrogate reduces the number of coupling variables and stored local distances from quadratic to linear in $n$. The runtime improvements from this smaller representation are evaluated empirically below.

{\subsection{Computational Results}
We compare accelerated and unaccelerated Marginal-DRO and MMD-DRO fits on matched synthetic classification datasets generated with \texttt{sklearn.datasets.make\_classification}. The accelerated MMD-DRO runs use 1,000 Nystr\"om components. We evaluate paired performance over 50 random seeds and report mean wall-clock times, median paired speedup ratios, and mean paired absolute differences in accuracy, $|\Delta A|$, on independent held-out datasets.

\input{dro_acceleration_tables.tex}
Tables~\ref{tab:marginal-dro-comparison} and~\ref{tab:mmd-dro-rank1000-comparison} report results across sample sizes, numbers of features, and DRO radii. Across these configurations, the accelerated versions achieve speedups of roughly 10--150$\times$, with greater computational gains tending to occur at larger sample sizes. The mean paired absolute accuracy difference is about 1--3 percentage points in most instances across the tested radii (e.g., 0.1--2 for Marginal-DRO and 0.0005--0.002 for MMD-DRO). These results suggest similar held-out predictive accuracy in most tested settings.
}

\wty{We provide further guidance on choosing the hyperparameters for the accelerated versions. Regardless of the formulation, whether an approximation is acceptable depends on the application's tolerance for differences in held-out performance. More specifically,
\begin{itemize}[leftmargin=*]
    \item For MMD-DRO, \texttt{n\_components} controls  approximation fidelity. In our sweep over 100 - 1000 components, smaller numbers of components substantially degraded predictive performance. We recommend increasing \texttt{n\_components} as the sample size and number of features increase, while checking the stability of held-out accuracy. Acceleration should be validated against an unaccelerated fit on a tractable subset. In our synthetic experiments, the accuracy difference was at most 2.54 percentage points for the tested MMD radii up to 0.002, but reached 8.40 points at radius 0.005. These values illustrate sensitivity to the radius under the tested kernel and data scale; they do not prescribe a universal range. In practical settings, we recommend scaling continuous features and, for regression, continuous targets where appropriate, then selecting the radius after fixing the kernel and its scale. We recommend first testing values of \texttt{n\_components} from 5\% to 20\% of a tractable subset's sample size and comparing accelerated and unaccelerated fits using the same training and held-out data. Increase \texttt{n\_components} if necessary.
    \item For Marginal-DRO, the accelerated implementation is a structural surrogate rather than an equivalent reformulation, and it has no additional tuning hyperparameters. We suggest using the accelerated fit as a starting point, then comparing it with an unaccelerated fit on a tractable subset at the intended parameter settings.
\end{itemize}}

\section{User Demo}
\label{app:demo}
We illustrate data loading, model fitting, and diagnostics using a synthetic dataset generated by the package. Further examples and complete API documentation are available at \url{https://python-dro.org}.
\begin{lstlisting}[language=Python]
# User example of chi-square DRO
from dro.linear_model import Chi2DRO
from dro.data.dataloader_classification import classification_basic
X, y = classification_basic(d = 2, num_samples = 100, radius = 2)
model = Chi2DRO(input_dim = 2, model_type = 'logistic')
model.update({'eps':1}) # update parameter
model.fit(X, y) # fit model
model.worst_distribution(X, y) # return the worst-case distribution
model.evaluate(X, y) # bias-corrected logistic loss estimate
\end{lstlisting}

\wty{\section{Detailed Comparison with Existing Software}\label{app:compare}
We provide a detailed comparison of \texttt{dro} with related software that addresses robust learning and optimization problems:
\begin{itemize}[noitemsep,leftmargin=*]
    \item \emph{Optimization modeling systems.}
    \texttt{ROME}, \texttt{RSOME}, and \texttt{ROC++} let users specify robust or stochastic
    optimization models and automatically construct solver-ready
    formulations~\citep{goh2011robust,chen2020robust,chen2023rsome,
    vayanos2022roc++}. These tools lack encapsulation of the DRO workflow in a unified interface and are difficult to integrate into modern ML workflows. In comparison, our package instead provides ready-to-fit estimators for its supported supervised-learning DRO problems.
    \item \emph{DRO learning implementations.}
    \texttt{skwdro} implements various Wasserstein DRO formulations under a PyTorch
    customization interface~\citep{vincent2024texttt}. Many recent DRO models for neural networks are accompanied by code. For example, 
    \texttt{fast-dro} implements efficient $\chi^2$- and CVaR-based
    training, and \texttt{group\_DRO} implements training under subpopulation shifts~\citep{levy2020large,
    sagawa2020distributionally}. Building on these relevant prior methods
    and implementations, our package brings supported Wasserstein,
    $f$-divergence, MMD, group-shift, and hybrid formulations into
    one unified interface with further computational acceleration.
    \item \emph{Adversarial-learning software.} The norm-ball form of our neural Wasserstein DRO uses attacks from
    \texttt{Torchattacks} within a supported DRO training procedure for generating adversarial examples \citep{kim2020torchattacks}, while \texttt{RobustBench} and
    \texttt{OODRobustBench} provide resources for evaluating
    adversarial robustness, including under distribution shifts
    \citep{croce2021robustbench,li2024oodrobustbench}. Besides adversarial learning, our package also exposes other DRO training objectives.
    \item \emph{Application-specific implementations.} Published DRO software also addresses applications in portfolio optimization
    in \texttt{skfolio}~\citep{nicolini2025skfolio}, fairness in
    survival analysis~\citep{hu2024fairnesssurvival}, cold-start
    recommendation~\citep{lin2024temporally}, and power-system
    dispatch~\citep{guo2018stochasticopf}. These implementations address their respective application objectives and constraints.
    Our package focuses on fitting and comparing a large number of DRO
    formulations for supervised regression and classification. The supported personalization interfaces allow users to express application-specific supervised-learning losses; additional decision variables and feasibility constraints require further modeling. 
\end{itemize}
}

\end{document}

%% file: dro_acceleration_tables.tex

\begin{table}[!htb]
\centering
\small
\caption{Marginal-DRO acceleration comparison, where the values following $\pm$ denote the standard errors of the means computed across 50 random seeds.}
\label{tab:marginal-dro-comparison}
\resizebox{\linewidth}{!}{%
\begin{tabular}{rrrrrrr}
\hline
Samples & Features & Radius & Base sec & Acc. sec & Acc. ratio & $|\Delta A|$ (pp) \\
\hline
1,000 & 50 & 0.1 & 13.317 {\scriptsize $\pm$ 0.065} & 0.263 {\scriptsize $\pm$ 0.026} & $55.9\times$ & 0.77 {\scriptsize $\pm$ 0.39} \\
1,000 & 50 & 0.5 & 13.303 {\scriptsize $\pm$ 0.064} & 0.2401 {\scriptsize $\pm$ 0.0057} & $57.0\times$ & 2.06 {\scriptsize $\pm$ 0.88} \\
1,000 & 50 & 1 & 13.262 {\scriptsize $\pm$ 0.070} & 0.2348 {\scriptsize $\pm$ 0.0027} & $56.7\times$ & 2.98 {\scriptsize $\pm$ 1.07} \\
1,000 & 50 & 2 & 13.270 {\scriptsize $\pm$ 0.069} & 0.2341 {\scriptsize $\pm$ 0.0027} & $56.9\times$ & 3.03 {\scriptsize $\pm$ 1.06} \\
\hline
1,000 & 100 & 0.1 & 13.91 {\scriptsize $\pm$ 0.15} & 0.465 {\scriptsize $\pm$ 0.017} & $31.0\times$ & 0.49 {\scriptsize $\pm$ 0.06} \\
1,000 & 100 & 0.5 & 13.83 {\scriptsize $\pm$ 0.15} & 0.4472 {\scriptsize $\pm$ 0.0065} & $31.0\times$ & 0.78 {\scriptsize $\pm$ 0.42} \\
1,000 & 100 & 1 & 13.82 {\scriptsize $\pm$ 0.14} & 0.4452 {\scriptsize $\pm$ 0.0064} & $30.7\times$ & 0.73 {\scriptsize $\pm$ 0.41} \\
1,000 & 100 & 2 & 13.77 {\scriptsize $\pm$ 0.13} & 0.4468 {\scriptsize $\pm$ 0.0064} & $30.8\times$ & 0.73 {\scriptsize $\pm$ 0.41} \\
\hline
5,000 & 50 & 0.1 & 451.0 {\scriptsize $\pm$ 3.4} & 3.181 {\scriptsize $\pm$ 0.058} & $142.6\times$ & 0.64 {\scriptsize $\pm$ 0.43} \\
5,000 & 50 & 0.5 & 450.2 {\scriptsize $\pm$ 3.2} & 3.108 {\scriptsize $\pm$ 0.051} & $146.1\times$ & 1.97 {\scriptsize $\pm$ 0.90} \\
5,000 & 50 & 1 & 449.4 {\scriptsize $\pm$ 3.2} & 3.079 {\scriptsize $\pm$ 0.048} & $144.1\times$ & 3.06 {\scriptsize $\pm$ 1.16} \\
5,000 & 50 & 2 & 451.6 {\scriptsize $\pm$ 3.3} & 3.073 {\scriptsize $\pm$ 0.048} & $148.8\times$ & 3.64 {\scriptsize $\pm$ 1.26} \\
\hline
5,000 & 100 & 0.1 & 443.1 {\scriptsize $\pm$ 4.4} & 7.70 {\scriptsize $\pm$ 0.11} & $57.0\times$ & 0.81 {\scriptsize $\pm$ 0.55} \\
5,000 & 100 & 0.5 & 443.1 {\scriptsize $\pm$ 4.3} & 7.57 {\scriptsize $\pm$ 0.11} & $57.5\times$ & 1.78 {\scriptsize $\pm$ 0.92} \\
5,000 & 100 & 1 & 440.7 {\scriptsize $\pm$ 4.0} & 7.53 {\scriptsize $\pm$ 0.11} & $57.5\times$ & 1.80 {\scriptsize $\pm$ 0.92} \\
5,000 & 100 & 2 & 441.5 {\scriptsize $\pm$ 4.3} & 7.46 {\scriptsize $\pm$ 0.11} & $57.6\times$ & 1.81 {\scriptsize $\pm$ 0.92} \\
\hline
\end{tabular}%
}
\end{table}

\begin{table}[!htb]
\centering
\small
\caption{MMD-DRO acceleration comparison with 1,000 Nystr\"om components, where the values following $\pm$ denote the standard errors of the means computed across 50 random seeds.}
\label{tab:mmd-dro-rank1000-comparison}
\resizebox{\linewidth}{!}{%
\begin{tabular}{rrrrrrr}
\hline
Samples & Features & Radius & Base sec & Acc. sec & Acc. ratio & $|\Delta A|$ (pp) \\
\hline
1,000 & 50 & 0.0005 & 1276 {\scriptsize $\pm$ 18} & 98.53 {\scriptsize $\pm$ 0.66} & $12.9\times$ & 0.54 {\scriptsize $\pm$ 0.08} \\
1,000 & 50 & 0.001 & 1252 {\scriptsize $\pm$ 25} & 94.01 {\scriptsize $\pm$ 0.84} & $13.0\times$ & 0.66 {\scriptsize $\pm$ 0.10} \\
1,000 & 50 & 0.002 & 1240 {\scriptsize $\pm$ 15} & 98.5 {\scriptsize $\pm$ 1.5} & $12.5\times$ & 0.70 {\scriptsize $\pm$ 0.10} \\
1,000 & 50 & 0.005 & 1389 {\scriptsize $\pm$ 33} & 95.50 {\scriptsize $\pm$ 0.64} & $14.4\times$ & 1.30 {\scriptsize $\pm$ 0.18} \\
\hline
1,000 & 100 & 0.0005 & 1333 {\scriptsize $\pm$ 27} & 106.0 {\scriptsize $\pm$ 1.3} & $12.3\times$ & 0.74 {\scriptsize $\pm$ 0.14} \\
1,000 & 100 & 0.001 & 1298 {\scriptsize $\pm$ 19} & 105.12 {\scriptsize $\pm$ 0.88} & $12.0\times$ & 0.82 {\scriptsize $\pm$ 0.16} \\
1,000 & 100 & 0.002 & 1293 {\scriptsize $\pm$ 16} & 102.04 {\scriptsize $\pm$ 0.77} & $12.5\times$ & 1.46 {\scriptsize $\pm$ 0.13} \\
1,000 & 100 & 0.005 & 1348 {\scriptsize $\pm$ 24} & 107.24 {\scriptsize $\pm$ 0.78} & $12.4\times$ & 3.84 {\scriptsize $\pm$ 1.07} \\
\hline
2,000 & 50 & 0.0005 & 13568 {\scriptsize $\pm$ 399} & 413.2 {\scriptsize $\pm$ 3.9} & $31.3\times$ & 1.38 {\scriptsize $\pm$ 0.20} \\
2,000 & 50 & 0.001 & 13051 {\scriptsize $\pm$ 339} & 421.1 {\scriptsize $\pm$ 2.1} & $30.5\times$ & 1.52 {\scriptsize $\pm$ 0.19} \\
2,000 & 50 & 0.002 & 13309 {\scriptsize $\pm$ 372} & 460.5 {\scriptsize $\pm$ 6.2} & $29.1\times$ & 1.63 {\scriptsize $\pm$ 0.20} \\
2,000 & 50 & 0.005 & 12860 {\scriptsize $\pm$ 399} & 399.9 {\scriptsize $\pm$ 3.4} & $30.3\times$ & 1.70 {\scriptsize $\pm$ 0.23} \\
\hline
2,000 & 100 & 0.0005 & 13407 {\scriptsize $\pm$ 231} & 464.8 {\scriptsize $\pm$ 3.3} & $28.4\times$ & 1.33 {\scriptsize $\pm$ 0.23} \\
2,000 & 100 & 0.001 & 13890 {\scriptsize $\pm$ 348} & 487 {\scriptsize $\pm$ 13} & $28.2\times$ & 1.90 {\scriptsize $\pm$ 0.23} \\
2,000 & 100 & 0.002 & 12527 {\scriptsize $\pm$ 251} & 456.0 {\scriptsize $\pm$ 3.3} & $26.7\times$ & 2.54 {\scriptsize $\pm$ 0.28} \\
2,000 & 100 & 0.005 & 13497 {\scriptsize $\pm$ 321} & 475.5 {\scriptsize $\pm$ 5.9} & $27.8\times$ & 8.40 {\scriptsize $\pm$ 1.74} \\
\hline
\end{tabular}%
}
\end{table}